\documentclass{article}%
\usepackage{amsmath}
\usepackage{amsfonts}
\usepackage{amssymb}
\usepackage{graphicx}
\usepackage{algorithmic}
\usepackage{algorithm}%
\providecommand{\U}[1]{\protect\rule{.1in}{.1in}}

\begin{document}

\title{SurvLIME-Inf: A simplified modification of SurvLIME for explanation of machine
learning survival models}
\author{Lev V. Utkin, Maxim S. Kovalev and Ernest M. Kasimov\\Peter the Great St.Petersburg Polytechnic University (SPbPU)\\St.Petersburg, Russia\\e-mail: lev.utkin@gmail.com, maxkovalev03@gmail.com, kasimov.ernest@gmail.com}
\date{}
\maketitle

\begin{abstract}
A new modification of the explanation method SurvLIME called SurvLIME-Inf for
explaining machine learning survival models is proposed. The basic idea behind
SurvLIME\ as well as SurvLIME-Inf is to apply the Cox proportional hazards
model to approximate the black-box survival model at the local area around a
test example. The Cox model is used due to the linear relationship of
covariates. In contrast to SurvLIME, the proposed modification uses
$L_{\infty}$-norm for defining distances between approximating and
approximated cumulative hazard functions. This leads to a simple linear
programming problem for determining important features and for explaining the
black-box model prediction. Moreover, SurvLIME-Inf outperforms SurvLIME when
the training set is very small. Numerical experiments with synthetic and real
datasets demonstrate the SurvLIME-Inf efficiency.


\end{abstract}

\section{Introduction}

Deep machine learning models can be regarded as powerful tools for solving
many applied problems, including medical diagnostics, finances, manufacturing
etc. In spite of satisfactory performance of the deep learning models, they
may be of limited use because their predictions are typically hard to be
interpreted or explained by human. This is caused by the fact that many
machine learning models work as black-box models. At the same time, there is a
high demand for understanding predictions produced by deep learning models.
For example, a doctor has to get an explanation of a diagnosis predicted by a
black-box model in order to choose a corresponding treatment. She or he has to
understand how the particular decisions are made by the deep learning model.
Therefore, explainability of deep learning models is an topical direction of
research nowadays, and, as a result, a lot of methods have been developed to
address the interpretation problems and to get accurate explanations for
obtained predictions
\cite{Arya-etal-2019,Buhrmester-etal-2019,Carvalho-etal-2019,Guidotti-2019,Holzinger-etal-2017,Mohseni-etal-2018,Molnar-2019,Murdoch-etal-2019,Tjoa-Guan-2019}%
.

There are two main groups of methods for explaining the black-box models. The
first group includes local methods which aim to interpret a single prediction.
The methods are based on using a local area around a test instance. The second
group consists of global methods. In contrast to the local methods, they
explain a black-box model on the whole input space or its part to take into
account the overall behavior of the model. We study only the local models
because our aim is to find features which lead to the individual prediction.
Moreover, we consider post-hoc explanation methods which are used to explain
predictions of such black-box models after they are trained.

One of the well-known local explanation methods is the Local Interpretable
Model-agnostic Explanations (LIME) \cite{Ribeiro-etal-2016}, which uses simple
and easily understandable linear models to locally approximate the predictions
of black-box models. LIME provides an explanation for a single instance by
perturbing it around its neighborhood and then by constructing a local
surrogate model trained also on original training data. Garreau and Luxburg
\cite{Garreau-Luxburg-2020} proposed a thorough theoretical analysis of the
LIME and derived closed-form expressions for coefficients of the explaining
model for the case of the linear explanation function. Garreau and Luxburg
\cite{Garreau-Luxburg-2020} point out that LIME is flexible to provide
explanations for different data types, including text and image data, while
being model-agnostic, i.e., any details of the black-box model are unknown.

One of the peculiarities of LIME is that it explains point-valued predictions
produced by the black-box model. However, there are models which produce
functions as predictions instead of points. The well-known class of the model
with these predictions is machine learning survival models
\cite{Lee-Zame-etal-2018,Zhao-Feng-2019} which solve survival analysis tasks
\cite{Hosmer-Lemeshow-May-2008,Wang-Li-Reddy-2017}.

One of the most widely-used regression models for the analysis of survival
data is the well-known Cox proportional hazards model, which is a
semi-parametric model that calculates effects of observed covariates on the
risk of an event occurring, for example, death or failure \cite{Cox-1972}. The
model assumes that a patient's log-risk of failure is a linear combination of
the instance covariates. This is a very important and at the same time very
strong assumption.

There are many survival analysis models, for example, random survival forests,
deep neural networks, etc., which relax this assumption and allow for more
general relationships between covariates and the output parameters
\cite{Wang-Li-Reddy-2017}. However, these models are the black-box ones and,
therefore, they require to be explained. Taking into account that predictions
of the models are functions, for example, the survival functions (SF),
cumulative hazard functions (CHF), the original LIME cannot be used. Kovalev
et al. \cite{Kovalev-Utkin-Kasimov-20} proposed an explanation method called
SurvLIME, which deals with censored data. The basic idea behind SurvLIME is to
apply the Cox model to approximate the black-box survival model at a local
area around a test instance. The Cox model is chosen due to its assumption of
the linear combination of covariates. Moreover, it is important that the
covariates as well as their combination do not depend on time. Therefore,
coefficients of the covariates can be regarded as quantitative impacts on the prediction.

SurvLIME includes a procedure which randomly generates synthetic instances
around the tested instance, and the CHF is calculated for every synthetic
instance by means of the black-box survival model. For every instance, the
approximating Cox model CHF is written as a function of coefficients of
interest. By writing the distance between CHFs provided by the black-box
survival model and by the approximating Cox model, respectively, an
unconstrained convex optimization problem for computing the coefficients of
covariates is constructed. The $L_{2}$-norm is used in order to consider the
distance between two CHF. As a result, the explanation by using SurvLIME is
based on solving the convex optimization problem. In order to simplify the
approach, we propose and investigate another explanation method which is based
on using $L_{\infty}$-norm for the distance between CHFs. This modification is
called SurvLIME-Inf.

The choice of this norm is caused by the fact that obtained optimization
problems become rather simple from the computational point of view
\cite{Marosevic-1996}. Indeed, we get the linear optimization problem for
computing coefficients of the Cox model. The $L_{\infty}$-norm (Chebychev
distance) is a measure of the approximation quality, which is defined as the
maximum of absolute values of the difference between the function being
approximated and the approximating function. Sim and Hartley
\cite{Sim-Hartley-2006} pointed out that $L_{\infty}$ minimization is not
robust to outliers, i.e., $L_{\infty}$ minimization may fit the outliers and
not the good data. Nevertheless, our experiments have illustrated a perfect
approximation of CHFs provided by the black-box survival model and the
approximating Cox model by rather small datasets.

Numerical results using synthetic and real data illustrate SurvLIME-Inf.

The paper is organized as follows. A short survey of publications devoted to
local explanation methods and machine learning models in survival analysis is
given in Section 2. Basic concepts of survival analysis are considered in
Section 3. A brief introduction to LIME can be found in Section 4. Basic ideas
behind SurvLIME-Inf are proposed in Section 5. Section 6 contains a formal
derivation of the linear programming problem implementing SurvLIME-Inf.
Numerical experiments with synthetic and real data are given in Section 7.
Concluding remarks are provided in Section 8.

\section{Related work}

\textbf{Local explanation methods.} LIME is one of the efficient and simple
explanation methods. As a result, many modifications of LIME have been
developed recently, including, DLIME \cite{Zafar-Khan-2019}, Anchor LIME
\cite{Ribeiro-etal-2018}, LIME-SUP \cite{Hu-Chen-Nair-Sudjianto-2018}, ALIME
\cite{Shankaranarayana-Runje-2019}, NormLIME \cite{Ahern-etal-2019},
LIME-Aleph \cite{Rabold-etal-2019}, GraphLIME \cite{Huang-Yamada-etal-2020},
MPS-LIME \cite{Shi-Zhang-Fan-20}, Tree-LIME \cite{Li-Fan-Shi-Chou-19},
SurvLIME \cite{Kovalev-Utkin-Kasimov-20}. Another popular method is the SHAP
\cite{Strumbel-Kononenko-2010} which takes a game-theoretic approach for
optimizing a regression loss function based on Shapley values
\cite{Lundberg-Lee-2017}. It is pointed out by Aas et al. \cite{Aas-etal-2019}
that Shapley values explain the difference between the prediction and the
global average prediction, while LIME explains the difference between the
prediction and a local average prediction.

Another group of explanation methods is based on counterfactual explanations
\cite{Wachter-etal-2017}, which try to explain what to do in order to achieve
a desired outcome by means of finding changes to some features of an
explainable input instance such that the resulting data point called
counterfactual has a different specified prediction than the original input.
It is important to note that LIME was also modified to implement
counterfactual explanations \cite{Ramon-etal-2020,White-Garcez-2020}.

Many explanation methods, including LIME, are based on perturbation techniques
\cite{Du-Liu-Hu-2019,Fong-Vedaldi-2019,Fong-Vedaldi-2017,Petsiuk-etal-2018,Vu-etal-2019}%
. These methods assume that contribution of a feature can be determined by
measuring how prediction score changes when the feature is altered.

Descriptions of many explanation methods and various approaches, their
critical reviews can be found in survey papers
\cite{Adadi-Berrada-2018,Arrieta-etal-2019,Carvalho-etal-2019,Guidotti-2019,Rudin-2019}%
.

Most methods explain point-valued predictions produced by black-box models,
i.e., predictions in the form of some number (class, regression value,
decision about anomaly, etc.). This fact restrict their use in survival
models, where predictions are usually represented in the form of CHFs or SFs.
Only SurvLIME \cite{Kovalev-Utkin-Kasimov-20} deals with these functions, but
it may be computationally hard due to the optimization problem which has to be solved.

\textbf{Machine learning models in survival analysis}. A clear taxonomy of
survival analysis methods and their comprehensive review can be found in
\cite{Wang-Li-Reddy-2017}. Following the Cox model \cite{Cox-1972}, a lot of
its modifications have been proposed. Some modifications are based on the
Lasso method \cite{Tibshirani-1997}, on the group Lasso penalty method
\cite{Kim-etal-2012}, on the adaptive Lasso \cite{Zhang-Lu-2007}. To relax
assumptions of the Cox model, in particular, the linear relationship between
covariates and the time of event, many models using neural networks, random
forests, support vector machines, etc. have been developed starting from the
pioneering work \cite{Faraggi-Simon-1995}. The corresponding review of the
methods is proposed by Nezhad et al. \cite{Nezhad-etal-2018}. One of the
important class of survival models, illustrating their efficiency and accuracy
especially by limited survival data, is the random survival forests (RSFs)
which can be viewed as an extension of the original random forests
\cite{Breiman-2001}. A detailed review of RSFs is presented by Bou-Hamad et
al. \cite{Bou-Hamad-etal-2011}.

Most survival models except for those based on the Cox model can be regarded
as black-box models. Therefore, they require to be explained in many
applications. At the same time, only the Cox model can be regarded as
explainable one due to its linear relationship between covariates. Therefore,
it will be used to approximate more powerful models, including survival deep
neural networks and RSFs, in order to explain predictions of these models.

\section{Basic concepts of survival analysis}

In survival analysis, an instance (patient) $i$ is represented by a triplet
$(\mathbf{x}_{i},\delta_{i},T_{i})$, where $\mathbf{x}_{i}^{\mathrm{T}%
}=(x_{i1},...,x_{id})$ is the vector of the patient parameters
(characteristics) or the vector of the instance features; $T_{i}$ is time to
event of the instance. If the event of interest is observed, then $T_{i}$
corresponds to the time between baseline time and the time of event happening,
in this case $\delta_{i}=1$, and we have an uncensored observation. If the
instance event is not observed and its time to event is greater than the
observation time, then $T_{i}$ corresponds to the time between baseline time
and end of the observation, and the event indicator is $\delta_{i}=0$, and we
have a censored observation. Suppose a training set $D$ consists of $n$
triplets $(\mathbf{x}_{i},\delta_{i},T_{i})$, $i=1,...,n$. The goal of
survival analysis is to estimate the time to the event of interest $T$ for a
new instance (patient) with a feature vector denoted by $\mathbf{x}$ by using
the training set $D$.

The survival and hazard functions are key concepts in survival analysis for
describing the distribution of event times. The survival function denoted by
$S(t|\mathbf{x})$ as a function of time $t$ is the probability of surviving up
to that time, i.e., $S(t|\mathbf{x})=\Pr\{T>t|\mathbf{x}\}$. The hazard
function $h(t|\mathbf{x})$ is defined as $h(t|\mathbf{x})=f(t|\mathbf{x}%
)/S(t|\mathbf{x})$, where $f(t|\mathbf{x})$ is the density function of the
event of interest.

Another important concept in survival analysis is the CHF $H(t|\mathbf{x})$,
which is defined as the integral of the hazard function $h(t|\mathbf{x})$ and
can be interpreted as the probability of an event at time $t$ given survival
until time $t$. The survival function is determined through the hazard
function and through the CHF as $S(t|\mathbf{x})=\exp\left(  -H(t|\mathbf{x}%
)\right)  $.

To compare survival models, the C-index proposed by Harrell et al.
\cite{Harrell-etal-1982} is used. It estimates how good a survival model is at
ranking survival times. In other words, this is the probability that the event
times of a pair of instances are correctly ranking. C-index does not depend on
choosing a fixed time for evaluation of the model and takes into account
censoring of patients \cite{May-etal-2004}.

According to the Cox proportional hazards model\cite{Cox-1972} , the hazard
function at time $t$ given predictor values $\mathbf{x}$ is defined as
\begin{equation}
h(t|\mathbf{x},\mathbf{b})=h_{0}(t)\exp\left(  \mathbf{b}^{\mathrm{T}%
}\mathbf{x}\right)  =h_{0}(t)\exp\left(  \sum\nolimits_{k=1}^{d}b_{k}%
x_{k}\right)  .
\end{equation}

Here $h_{0}(t)$ is a baseline hazard function which does not depend on the
vector $\mathbf{x}$ and the vector $\mathbf{b}$; $\mathbf{b}^{\mathrm{T}%
}=(b_{1},...,b_{d})$ is an unknown vector of regression coefficients or parameters.

In the framework of the Cox model, the survival function $S(t|\mathbf{x}%
,\mathbf{b})$ is computed as
\begin{equation}
S(t|\mathbf{x},\mathbf{b})=\exp(-H_{0}(t)\exp\left(  \mathbf{b}^{\mathrm{T}%
}\mathbf{x}\right)  =\left(  S_{0}(t)\right)  ^{\exp\left(  \mathbf{b}%
^{\mathrm{T}}\mathbf{x}\right)  }.
\end{equation}

Here $H_{0}(t)$ is the cumulative baseline hazard function; $S_{0}(t)$ is the
baseline survival function. It is important to note that functions $H_{0}(t)$
and $S_{0}(t)$ do not depend on $\mathbf{x}$ and $\mathbf{b}$.

One of the ways for estimating parameters $\mathbf{b}$ of the Cox model is the
Cox partial likelihood function \cite{Cox-1972}. There are other methods,
including the Breslow approximation \cite{Breslow-1974} and the Efron
approximation \cite{Efron-1977}.

\section{LIME}

Before studying the LIME modification for survival data, this method is
briefly considered below. LIME proposes to approximate a black-box model
denoted as $f$ with a simple function $g$ in the vicinity of the point of
interest $\mathbf{x}$, whose prediction by means of $f$ has to be explained,
under condition that the approximation function $g$ belongs to a set of
explanation models $G$, for example, linear models. In order to construct the
function $g$ in accordance with LIME, a new dataset consisting of perturbed
samples is generated, and predictions corresponding to the perturbed samples
are obtained by means of the explained model. New samples are assigned by
weights $w_{\mathbf{x}}$ in accordance with their proximity to the point of
interest $\mathbf{x}$ by using a distance metric, for example, the Euclidean
distance or a kernel. The weights are used to enforce locality for the linear
model $g$.

An explanation (local surrogate) model is trained on new generated samples by
solving the following optimization problem:
\begin{equation}
\arg\min_{g\in G}L(f,g,w_{\mathbf{x}})+\Phi(g).
\end{equation}

Here $L$ is a loss function, for example, mean squared error, which measures
how the explanation is close to the prediction of the black-box model;
$\Phi(g)$ is the model complexity.

As a result, the prediction is explained by analyzing coefficients of the
local linear model. The output of LIME, therefore, is a set of important
features corresponding to coefficients of the linear model.

\section{A general algorithm of SurvLIME and SurvLIME-Inf}

Suppose that there are a training set $D$ and a black-box model which produces
an output in the form of the CHF $H(t|\mathbf{x})$ for every new instance
$\mathbf{x}$. An idea behind SurvLIME is to approximate the output of the
black-box model with the CHF produced by the Cox model for the same input
instance $\mathbf{x}$. With this approximation, we get the parameters
$\mathbf{b}$ of the approximating Cox model, whose values can be regarded as
quantitative impacts on the prediction $H(t|\mathbf{x})$. The largest
coefficients indicate the corresponding important features.

Denote the Cox CHF as $H_{\text{Cox}}(t|\mathbf{x},\mathbf{b})$. Then we have
to find such parameters $\mathbf{b}$ that the distance between $H(t|\mathbf{x}%
)$ and $H_{\text{Cox}}(t|\mathbf{x},\mathbf{b})$ for the considered instance
$\mathbf{x}$ would be as small as possible. In order to avoid incorrect
results, a lot of nearest points $\mathbf{x}_{k}$ in a local area around
$\mathbf{x}$ is generated. For every $\mathbf{x}_{k}$, the CHF $H(t|\mathbf{x}%
_{k})$ of the black-box model is obtained as a prediction of the black-box
model. Now optimal values of $\mathbf{b}$ can be computed by minimizing the
weighted average distance between every pair of CHFs $H(t|\mathbf{x}_{k})$ and
$H_{\text{Cox}}(t|\mathbf{x}_{k},\mathbf{b})$ over all points $\mathbf{x}_{k}%
$. Weight $w_{k}$ assigned to the $k$-th distance depends on the distance
between $\mathbf{x}_{k}$ and $\mathbf{x}$. Smaller distances between
$\mathbf{x}_{k}$ and $\mathbf{x}$ produce larger weights of distances between CHFs.

It is important to point out that the optimization problem for computing
parameters $\mathbf{b}$ depends on the used distance metric between CHFs
$H(t|\mathbf{x}_{k})$ and $H_{\text{Cox}}(t|\mathbf{x}_{k},\mathbf{b})$.
SurvLIME uses the $L_{2}$-norm which leads to a convex optimization problem.
SurvLIME-Inf uses the $L_{\infty}$-norm. We will show that this distance
metric leads to the linear programming problem whose solution is very simple.

\section{Optimization problem for computing parameters}

Let $t_{0}<t_{1}<...<t_{m}$ be the distinct times to event of interest, for
example, times to deaths from the set $\{T_{1},...,T_{n}\}$, where $t_{0}%
=\min_{k=1,...,n}T_{k}$ and $t_{m}=\max_{k=1,...,n}T_{k}$. The black-box model
maps the feature vectors $\mathbf{x}\in\mathbb{R}^{d}$ into piecewise constant
CHFs $H(t|\mathbf{x})$ such that $H(t|\mathbf{x})\geq0$ for all $t$, $\max
_{t}H(t|\mathbf{x})<\infty$. Let us introduce the time $T\geq t_{m}$ in order
to restrict $H(t|\mathbf{x})$ and denote $\Omega=[0,T]$.

Interval $\Omega$ can be divided into into $m+1$ non-intersecting subsets
$\Omega_{0},...,\Omega_{m}$ such that $\Omega_{j}=[t_{j},t_{j+1})$, $\forall
j\in\{0,...,m-1\}$, $\Omega_{m}=[t_{m},T]$. After introducing the indicator
functions $I_{j}(t)$, which takes the value $1$ when $t\in\Omega_{j}$, and $0$
otherwise, we rewrite $H(t|\mathbf{x})$ as follows:
\begin{equation}
H(t|\mathbf{x})=\sum_{j=0}^{m}H_{j}(\mathbf{x})\cdot I_{j}(t).
\end{equation}

Here $H_{j}(\mathbf{x})$ is a part of $H(t|\mathbf{x})$ in interval
$\Omega_{j}$ under additional condition $H_{j}(\mathbf{x})>0$. CHF
$H_{j}(\mathbf{x})$ does not depend on $t$ in interval $\Omega_{j}$ because it
is constant in this interval.

The same can be written for the Cox CHF:
\begin{equation}
H_{\text{Cox}}(t|\mathbf{x},\mathbf{b})=H_{0}(t)\exp\left(  \mathbf{b}%
^{\mathrm{T}}\mathbf{x}\right)  =\sum_{j=0}^{m}\left[  H_{0j}\exp\left(
\mathbf{b}^{\mathrm{T}}\mathbf{x}\right)  \right]  \cdot I_{j}(t).
\end{equation}

It should be noted that the use CHFs for computing the distance between them
leads to a complex optimization problem which may be non-convex. Therefore, we
proposed to take logarithms of $H(t|\mathbf{x})$ and $H_{\text{Cox}%
}(t|\mathbf{x},\mathbf{b})$ denoted as $\phi(t|\mathbf{x})$ and $\phi
_{\text{Cox}}(t|\mathbf{x},\mathbf{b})$, respectively. Since the logarithm is
the monotone function, then there hold%

\begin{equation}
\phi(t|\mathbf{x})=\sum_{j=0}^{m}(\ln H_{j}(\mathbf{x}))I_{j}(t),
\label{SurvLIME-Inf-32}%
\end{equation}%
\begin{align}
\phi_{\text{Cox}}(t|\mathbf{x},\mathbf{b})  &  =\sum_{j=0}^{m}\left(
\ln\left[  H_{0j}\exp\left(  \mathbf{b}^{\mathrm{T}}\mathbf{x}\right)
\right]  \right)  I_{j}(t)\nonumber\\
&  =\sum_{j=0}^{m}\left(  \ln H_{j}(\mathbf{x})-\ln H_{0j}-\mathbf{b}%
^{\mathrm{T}}\mathbf{x}\right)  I_{j}(t). \label{SurvLIME-Inf-33}%
\end{align}

Let us consider the distance between $\phi(t|\mathbf{x}_{k})$ and
$\phi_{\text{Cox}}(t|\mathbf{x}_{k},\mathbf{b})$ based on the $L_{\infty}%
$-norm for every generated point $\mathbf{x}_{k}$:%
\begin{align}
D_{\infty,k}\left(  \phi,\phi_{\text{Cox}}\right)   &  =\left\Vert
\phi(t|\mathbf{x}_{k})-\phi_{\text{Cox}}(t|\mathbf{x}_{k},\mathbf{b}%
)\right\Vert _{\infty}\nonumber\\
&  =\max_{t\in\Omega}\left\vert \phi(t|\mathbf{x}_{k})-\phi_{\text{Cox}%
}(t|\mathbf{x}_{k},\mathbf{b})\right\vert .
\end{align}

Hence, the weighted average distance between $\phi(t|\mathbf{x}_{k})$ and
$\phi_{\text{Cox}}(t|\mathbf{x}_{k},\mathbf{b})$ for $N$ generated points
$\mathbf{x}_{k}$ has to be minimized over $\mathbf{b}$. This can be written as
the following optimization problem:
\begin{equation}
\min_{\mathbf{b}}\left(  \sum_{k=1}^{N}w_{k}\cdot\max_{t\in\Omega}\left\vert
\phi(t|\mathbf{x}_{k})-\phi_{\text{Cox}}(t|\mathbf{x},\mathbf{b})\right\vert
\right)  . \label{SurvLIME-Inf-40}%
\end{equation}

Let us introduce the optimization variables%
\begin{equation}
z_{k}=\max_{t\in\Omega}\left\vert \phi(t|\mathbf{x}_{k})-\phi_{\text{Cox}%
}(t|\mathbf{x}_{k},\mathbf{b})\right\vert .
\end{equation}

They are restricted as follows:
\begin{equation}
z_{k}\geq\left\vert \phi(t|\mathbf{x}_{k})-\phi_{\text{Cox}}(t|\mathbf{x}%
_{k},\mathbf{b})\right\vert ,\ \forall t\in\Omega.
\end{equation}
The above constraints for every $t$ can be represented as two constraints
\begin{equation}
z_{k}\geq\phi(t|\mathbf{x}_{k})-\phi_{\text{Cox}}(t|\mathbf{x}_{k}%
,\mathbf{b}),\ \forall t\in\Omega, \label{SurvLIME-Inf-42}%
\end{equation}%
\begin{equation}
z_{k}\geq\phi_{\text{Cox}}(t|\mathbf{x}_{k},\mathbf{b})-\phi(t|\mathbf{x}%
_{k}),\ \forall t\in\Omega. \label{SurvLIME-Inf-43}%
\end{equation}

Substituting (\ref{SurvLIME-Inf-32})-(\ref{SurvLIME-Inf-33}) into
(\ref{SurvLIME-Inf-40}) and taking into account (\ref{SurvLIME-Inf-42}%
)-(\ref{SurvLIME-Inf-43}), we get%
\begin{equation}
\min_{\mathbf{b}}\sum_{k=1}^{N}w_{k}z_{k},
\end{equation}
subject to $\forall t\in\Omega$ and $k=1,...,N$,
\begin{equation}
z_{k}\geq\sum_{j=0}^{m}\left(  \ln H_{j}(\mathbf{x}_{k})-\ln H_{0j}%
-\mathbf{b}^{\mathrm{T}}\mathbf{x}_{k}\right)  I_{j}(t),
\end{equation}%
\begin{equation}
z_{k}\geq\sum_{j=0}^{m}\left(  \ln H_{0j}+\mathbf{b}^{\mathrm{T}}%
\mathbf{x}_{k}-\ln H_{j}(\mathbf{x}_{k})\right)  I_{j}(t).
\end{equation}

The last constraints can be rewritten as
\begin{equation}
z_{k}\geq\ln H_{j}(\mathbf{x}_{k})-\ln H_{0j}-\mathbf{b}^{\mathrm{T}%
}\mathbf{x}_{k},\ j=0,...,m,
\end{equation}%
\begin{equation}
z_{k}\geq\mathbf{b}^{\mathrm{T}}\mathbf{x}_{k}\mathbf{+}\ln H_{0j}-\ln
H_{j}(\mathbf{x}_{k}),\ j=0,...,m.
\end{equation}

Note that term $\mathbf{b}^{\mathrm{T}}\mathbf{x}_{k}$ does not depend on $j$.
This implies that the constraints can be reduced to the following simple
constraints:
\begin{equation}
z_{k}\geq Q_{k}-\mathbf{b}^{\mathrm{T}}\mathbf{x},\ k=1,...,N,
\end{equation}%
\begin{equation}
z_{k}\geq\mathbf{b}^{\mathrm{T}}\mathbf{x}-R_{k},\ k=1,...,N,
\end{equation}
where%
\begin{equation}
Q_{k}=\max_{j=0,...,m}\left(  \ln H_{j}(\mathbf{x}_{k})-\ln H_{0j}\right)  ,
\end{equation}%
\begin{equation}
R_{k}=\min_{j=0,...,m}\left(  \ln H_{j}(\mathbf{x}_{k})-\ln H_{0j}\right)  .
\end{equation}

Finally, we get the linear optimization problem with $d+N$ optimization
variables ($z_{1},...,z_{N}$ and $\mathbf{b}$) and $2N$ constraints. It is of
the form:
\begin{equation}
\min_{\mathbf{b}}\sum_{k=1}^{N}w_{k}z_{k}, \label{SurvLIME-Inf-60}%
\end{equation}
subject to%
\begin{equation}
z_{k}\geq Q_{k}-\mathbf{x}_{k}\mathbf{b}^{\mathrm{T}},\ k=1,...,N,
\label{SurvLIME-Inf-61}%
\end{equation}%
\begin{equation}
z_{k}\geq\mathbf{x}_{k}\mathbf{b}^{\mathrm{T}}-R_{k},\ k=1,...,N.
\label{SurvLIME-Inf-62}%
\end{equation}

Finally, we write the following scheme of Algorithm \ref{alg:SurvLIME-Infin}.

\begin{algorithm}
\caption{The algorithm for computing vector $\bf {b}$ for point $\bf {x}$ in SurvLIME-Inf}\label{alg:SurvLIME-Infin}%

\begin{algorithmic}
[1]\REQUIRE Training set $D$; point of interest $\mathbf{x}$; the number of
generated points $N$; the black-box survival model for explaining
$f(\mathbf{x})$ \ENSURE Vector $\mathbf{b}$ of important features \STATE
Compute the baseline CHF $H_{0}(t)$ of the approximating Cox model on dataset
$D$ by using the Nelson--Aalen estimator

\STATE Generate $N-1$ random nearest points $\mathbf{x}_{k}$ in a local area
around $\mathbf{x}$, point $\mathbf{x}$ is the $N$-th point

\STATE Get the prediction of $H(t|\mathbf{x}_{k})$ by using the black-box
survival model (the function $f$)

\STATE Compute weights $w_{k}=K(\mathbf{x},\mathbf{x}_{k})$ of perturbed
points, $k=1,...,N$

\STATE Find vector $\mathbf{b}$ by solving the convex optimization problem
(\ref{SurvLIME-Inf-60})-(\ref{SurvLIME-Inf-62})
\end{algorithmic}
\end{algorithm}

\section{Numerical experiments}

\subsection{Synthetic data}

In order to investigate the proposed method, random survival times to events
are generated by using the Cox model estimates. For experiments, we randomly
generate $N=1000$ covariate vectors $\mathbf{x}\in\mathbb{R}^{d}$ from the
uniform distribution in the $d$-sphere with predefined radius $R=8$. Here
$d=5$. The center of the sphere is $p=(0,0,0,0,0)$. There are several methods
for the uniform sampling of points $\mathbf{x}$ in the $d$-sphere with the
unit radius $R=1$, for example, \cite{Barthe-etal-2005,Harman-Lacko-2010}.
Then every generated point is multiplied by $R$.

We use the Cox model estimates to generate random survival times, applying
results obtained by Bender et al. \cite{Bender-etal-2005} for survival time
data for the Cox model with Weibull distributed survival times. The Weibull
distribution with the scale $\lambda=10^{-5}$ and shape $v=2$ parameters is
used to generate appropriate survival times for simulation studies because
this distribution shares the assumption of proportional hazards with the Cox
regression model \cite{Bender-etal-2005}. If we take the vector $\mathbf{b}%
^{\mathrm{T}}=(-0.25,10^{-6},-0.1,0.35,10^{-6})$, then the following
expression can be used for generating survival times \cite{Bender-etal-2005}:
\begin{equation}
T=\left(  \frac{-\ln(U)}{\lambda\exp(\mathbf{b}^{\mathrm{T}}\mathbf{x}%
)}\right)  ^{1/v}, \label{SurvLIME1_84}%
\end{equation}
where $U$ is the random variable uniformly distributed in interval $[0,1]$.

It can be seen that vector $\mathbf{b}$ has two almost zero-valued elements
and three \textquotedblleft large\textquotedblright\ elements which will
correspond to important features. Generated values $T_{i}$ are restricted by
the condition: if $T_{i}>2000$, then $T_{i}$ is replaced with value $2000$.
The event indicator $\delta_{i}$ is generated from the binomial distribution
with probabilities $\Pr\{\delta_{i}=1\}=0.9$, $\Pr\{\delta_{i}=0\}=0.1$.

Perturbations can be viewed as a step of the algorithm. According to it, $N$
nearest points $\mathbf{x}_{k}$ are generated in a local area around
$\mathbf{x}$. These points are uniformly generated in the $d$-sphere with some
predefined radius $r=0.5$ and with the center at point $\mathbf{x}$. Weights
to every point are assigned as follows:
\begin{equation}
w_{k}=1-\left(  r^{-1}\cdot\left\Vert \mathbf{x}-\mathbf{x}_{k}\right\Vert
_{2}\right)  ^{1/2}. \label{SurvLIME_60}%
\end{equation}

To compare vectors $\mathbf{b}$, we introduce the following notation:
$\mathbf{b}^{\text{model}}$ are coefficients of the Cox model which is used as
the black-box model; $\mathbf{b}^{\text{true}}$ are coefficients used for
training data generation (see (\ref{SurvLIME1_84})); $\mathbf{b}^{\text{expl}%
}$ are explaining coefficients obtained by using the proposed explanation algorithm.

One of the aims of numerical experiments is to consider the method behavior by
assuming that the black-box model is the Cox model. With these experiments, we
have an opportunity to compare the vector $\mathbf{b}^{\text{true}}$ with
vectors $\mathbf{b}^{\text{model}}$ and $\mathbf{b}^{\text{expl}}$ because the
black-box Cox model as well as the explanation Cox model are expected to have
close vectors $\mathbf{b}$. We cannot perform the same comparison by using the
RSF as a black-box model. Therefore, the results with the RSF will be compared
by considering the proximity of SFs obtained from the RSF and the explanation
Cox model.

To evaluate the algorithm, 900 instances are randomly selected from every
cluster for training and 100 instances are for testing. In the test phase, the
optimal explanation vector $\mathbf{b}^{\text{expl}}$ is computed for every
point from the testing set. In accordance with the obtained vectors
$\mathbf{b}^{\text{expl}}$, we depict the best, mean and worst approximations
on the basis of the Euclidean distance between vectors $\mathbf{b}%
^{\text{expl}}$ and $\mathbf{b}^{\text{model}}$ (for the Cox model) and with
Euclidean distance between $H(t_{j}|\mathbf{x}_{i})$ and $H_{\text{Cox}%
}\left(  t_{j}|\mathbf{x}_{i},\mathbf{b}_{i}^{\text{expl}}\right)  $ (for the
RSF). In order to get these approximations, points with the best, mean and
worst approximations are selected among all testing points.

\begin{figure}
[ptb]
\begin{center}
\includegraphics[
height=3.7438in,
width=3.8951in
]%
{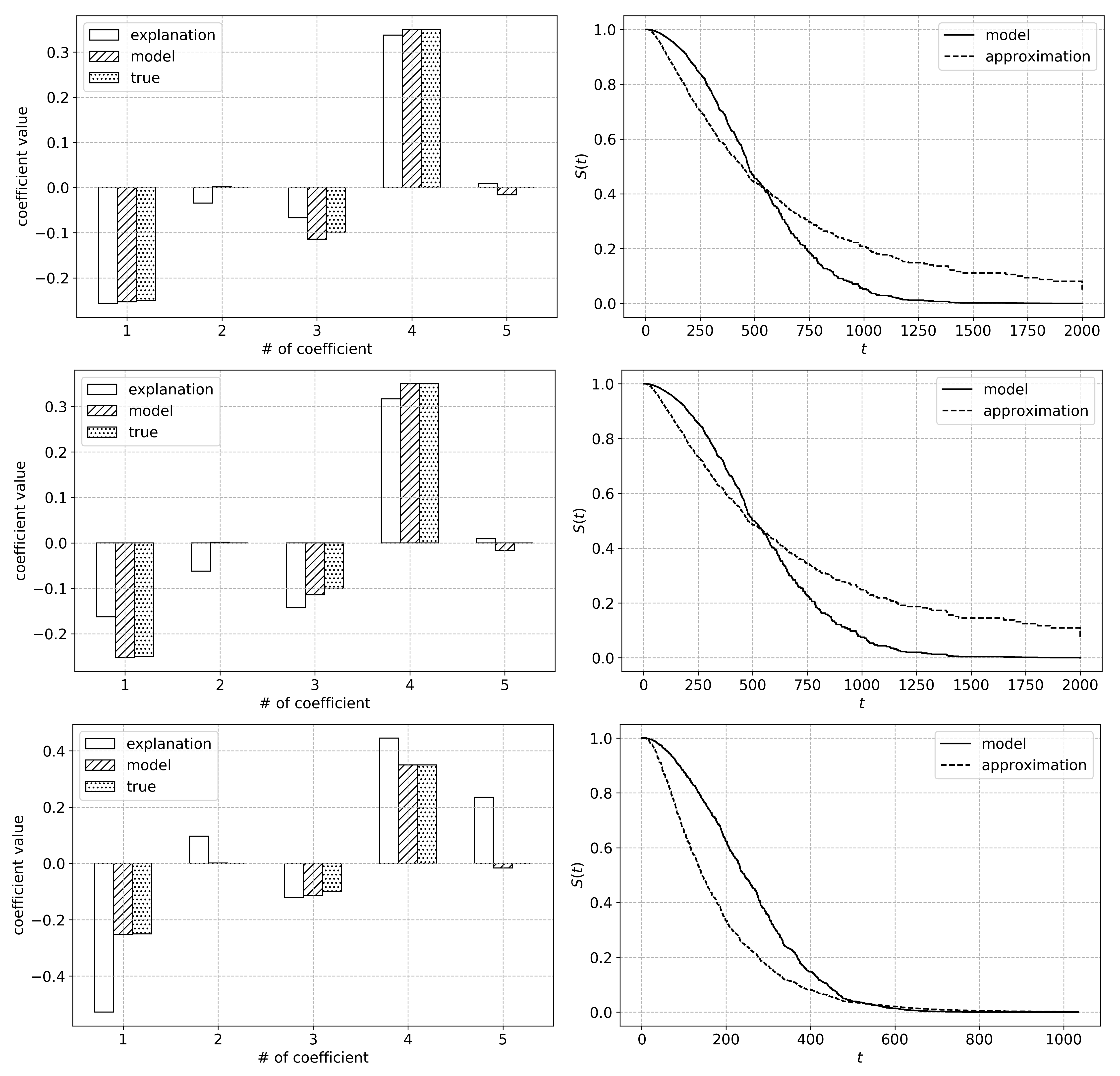}%
\caption{The best, mean and worst approximations for the Cox model}%
\label{f:cox_synth}%
\end{center}
\end{figure}

The three cases (best (pictures in the first row), mean (pictures in the
second row) and worst (pictures in the third row)) of approximations for the
black-box Cox model are depicted in Fig. \ref{f:cox_synth}. Left pictures show
values of important features $\mathbf{b}^{\text{expl}}$, $\mathbf{b}%
^{\text{model}}$ and $\mathbf{b}^{\text{true}}$. It can be seen from these
pictures that all experiments show very clear coincidence of important
features for all models. Right pictures in Fig. \ref{f:cox_synth} show SFs
computed by using the black-box Cox model and the Cox approximation. It
follows from the pictures that the approximation is perfect even for the worst case.%

\begin{figure}
[ptb]
\begin{center}
\includegraphics[
height=1.5082in,
width=5.4544in
]%
{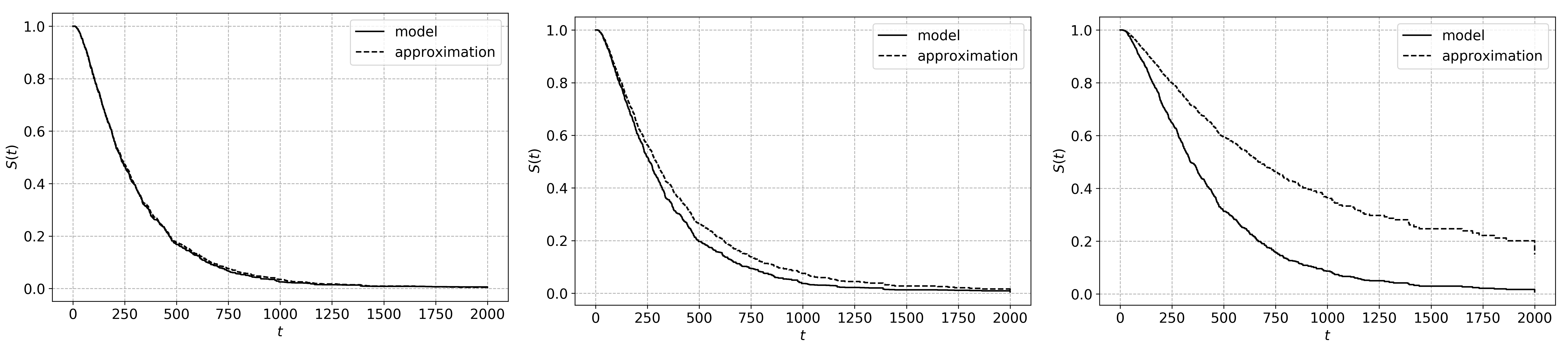}%
\caption{The best, mean and worst approximations for the RSF model}%
\label{f:rsf_synth}%
\end{center}
\end{figure}

Similar results for the black-box RSF model are shown in Fig.
\ref{f:rsf_synth} where three pictures correspond to the best, mean and worst
approximations. The important features are not shown in Fig. \ref{f:rsf_synth}
because RSF does not provide the important features like the Cox model.
However, it follows from the SFs in Fig. \ref{f:rsf_synth} that the proposed
method provides the perfect approximation of the RSF output by the Cox model.%

\begin{figure}
[ptb]
\begin{center}
\includegraphics[
height=4.4175in,
width=3.346in
]%
{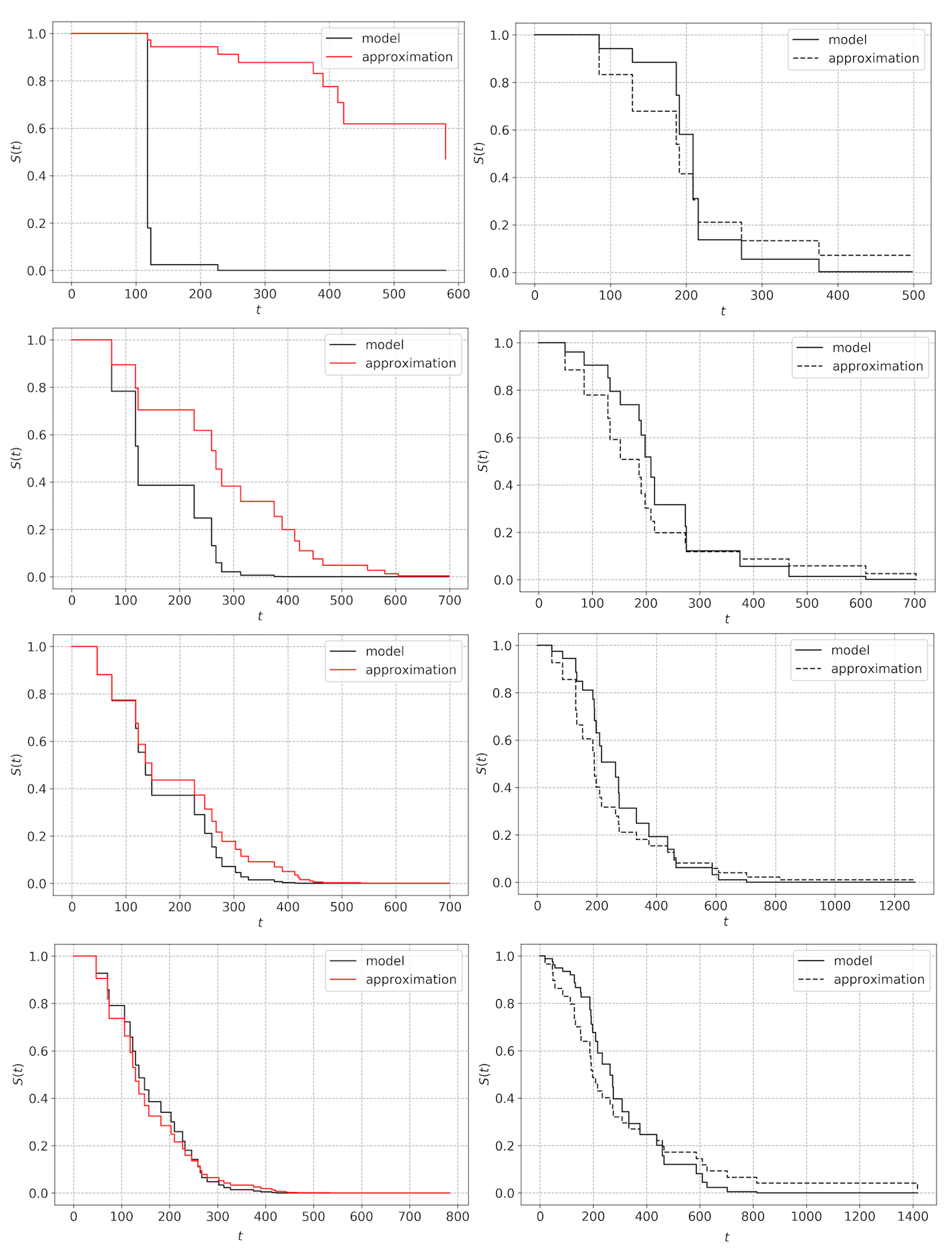}%
\caption{Comparison of the SFs relationship for SurvLIME and SurvLIME-Inf with
using the black-box Cox model by $10$, $20$, $30$, $40$ training examples}%
\label{fig:cox_10_inf}%
\end{center}
\end{figure}

It is interesting to point out that SurvLIME-Inf provides better results in
comparison with SurvLIME for small amounts of training data. In order to study
the method under this condition, the black-box Cox model and the RSF are
trained on $10$, $20$, $30$, $40$ examples. The models are tested on $10$
examples. Numerical results for the Cox model are shown in Fig.
\ref{fig:cox_10_inf} where rows correspond to $10$, $20$, $30$, $40$ training
examples, respectively, left pictures in every row show relationships between
SFs obtained from the black-box Cox model and from the Cox approximation by
using SurvLIME, the same relationships by using SurvLIME-Inf are depicted in
right pictures. One can see from Fig. \ref{fig:cox_10_inf} that SurvLIME-Inf
provides better approximations of SFs in comparison with SurvLIME for cases of
$10$, $20$, $30$ training examples. However, it can be seen from the last row
($40$ training examples) that SurvLIME becomes better with increase of the
training set.%

Measures $RMSE_{\text{model}}$ and $RMSE_{\text{true}}$ as functions of the
sample size $n$ for SurvLIME and SurvLIME-Inf are provided in Table
\ref{t:cox_10_inf} for comparison purposes. They are defined for the Cox model
from $n_{\text{test}}$ testing results as follows:%
\begin{equation}
RMSE_{\text{type}}=\left(  \frac{1}{n_{\text{test}}}\sum_{i=1}^{n_{\text{test}%
}}\left\Vert \mathbf{b}_{i}^{\text{type}}-\mathbf{b}_{i}^{\text{expl}%
}\right\Vert _{2}\right)  ^{1/2},
\end{equation}
where is \textquotedblleft model\textquotedblright\ and \textquotedblleft
true\textquotedblright\ is substituted into the above expression in place of
\textquotedblleft type\textquotedblright.

$RMSE_{\text{model}}$ characterizes how the obtained important features
coincide with the corresponding features obtained by using the Cox model as
the black-box model. $RMSE_{\text{true}}$ considers how the obtained important
features coincide with the features used for generating the random times to events.

It can be seen from Table \ref{t:cox_10_inf} that SurvLIME-Inf outperforms
SurvLIME for small $n$, namely, for $n=10$ and $20$. At the same time, this
outperformance disappears with increasing $n$, i.e., when $n=30$ and $40$.
This is a very interesting observation which tells us that SurvLIME-Inf should
be used when the training set is very small.%

\begin{table}[tbp] \centering
\caption{Approximation measures for four cases of using the black-box Cox model by the small amount of data for SurvLIME and SurvLIME-Inf}%
\begin{tabular}
[c]{ccccc}\hline
& \multicolumn{2}{c}{SurvLIME} & \multicolumn{2}{c}{SurvLIME-Inf}\\\hline
$n$ & $RMSE_{\text{model}}$ & $RMSE_{\text{true}}$ & $RMSE_{\text{model}}$ &
$RMSE_{\text{true}}$\\\hline
$10$ & $0.719$ & $0.809$ & $0.290$ & $0.575$\\\hline
$20$ & $0.659$ & $0.664$ & $0.358$ & $0.460$\\\hline
$30$ & $0.347$ & $0.428$ & $0.398$ & $0.432$\\\hline
$40$ & $0.324$ & $0.344$ & $0.388$ & $0.451$\\\hline
\end{tabular}
\label{t:cox_10_inf}%
\end{table}%

\begin{figure}
[ptb]
\begin{center}
\includegraphics[
height=4.4901in,
width=3.3434in
]%
{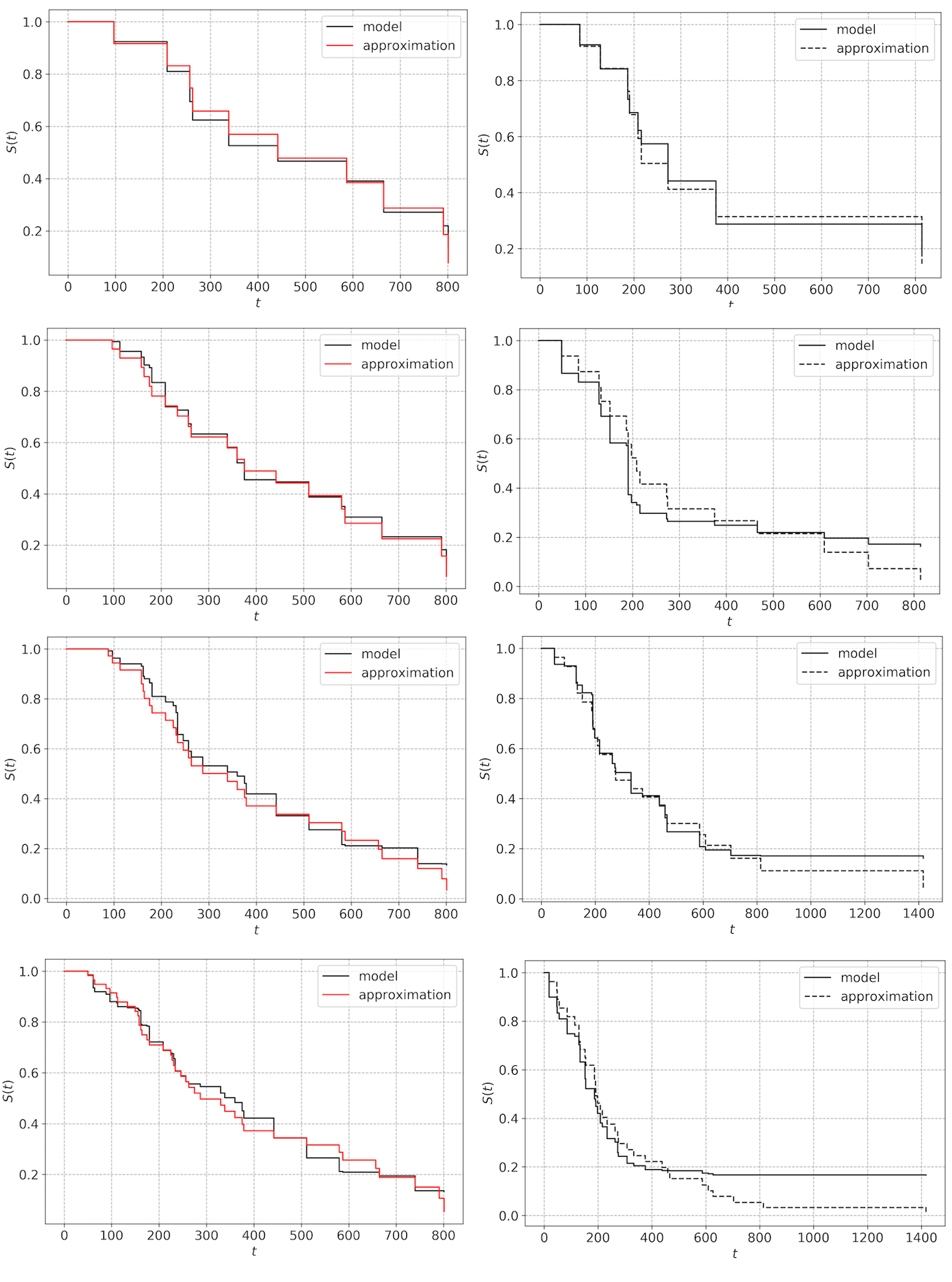}%
\caption{Comparison of the SFs relationship for SurvLIME and SurvLIME-Inf with
using the RSF by $10$, $20$, $30$, $40$ training examples}%
\label{fig:rsf_10_inf}%
\end{center}
\end{figure}

The same experiments are carried out for the RSF. They are shown in Fig.
\ref{fig:rsf_10_inf}.%

\subsection{Real data}

Let us apply the following well-known real datasets to study the method.

The Veterans' Administration Lung Cancer Study (Veteran) Dataset
\cite{Kalbfleisch-Prentice-1980} contains data on 137 males with advanced
inoperable lung cancer. The subjects were randomly assigned to either a
standard chemotherapy treatment or a test chemotherapy treatment. Several
additional variables were also measured on the subjects. The number of
features is 6, but it is extended till 9 due to categorical features.

The NCCTG Lung Cancer (LUNG) Dataset \cite{Loprinzi-etal-1994} records the
survival of patients with advanced lung cancer, together with assessments of
the patients performance status measured either by the physician and by the
patients themselves. The data set contains 228 patients, including 63 patients
that are right censored (patients that left the study before their death). The
number of features is 8, but it is extended till 11 due to categorical features.

The Primary Biliary Cirrhosis (PBC) Dataset contains observations of 418
patients with primary biliary cirrhosis of the liver from the Mayo Clinic
trial \cite{Fleming-Harrington-1991}, 257 of whom have censored data. Every
example is characterized by 17 features including age, sex, ascites, hepatom,
spiders, edema, bili and chol, etc. The number of features is extended till 22
due to categorical features.

The above datasets can be downloaded via the \textquotedblleft
survival\textquotedblright\ R package.

\begin{figure}
[ptb]
\begin{center}
\includegraphics[
height=1.6405in,
width=5.514in
]%
{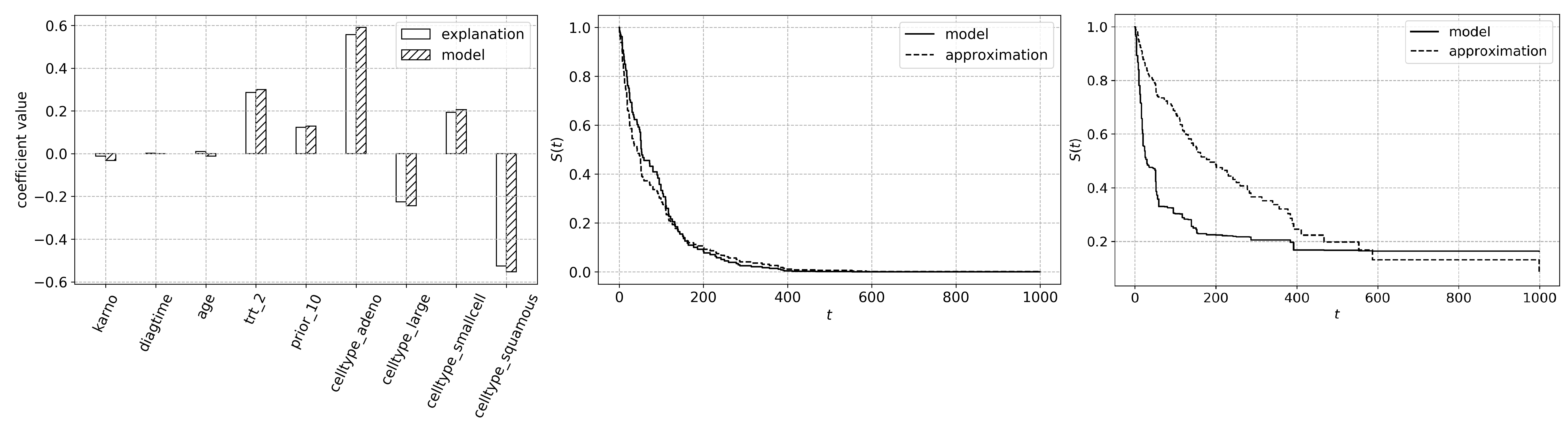}%
\caption{The mean approximation for the Cox model (the first and the second
picture) and the RSF (the third picture) trained on the Veteran dataset}%
\label{f:veteran_cox_rsf}%
\end{center}
\end{figure}

Fig. \ref{f:veteran_cox_rsf} illustrates numerical results for the Veteran
dataset. We provide only the case of the mean approximation in order to reduce
the number of similar pictures. Fig. \ref{f:veteran_cox_rsf} contains three
pictures: the first one illustrates the explanation important features and
important features computed by using the Cox model; the second picture shows
two SFs for the Cox model; the third picture shows two SFs for the RSF. It
follows from Fig. \ref{f:veteran_cox_rsf} that the method provides appropriate
results for the real dataset.

\begin{figure}
[ptb]
\begin{center}
\includegraphics[
height=1.465in,
width=5.5815in
]%
{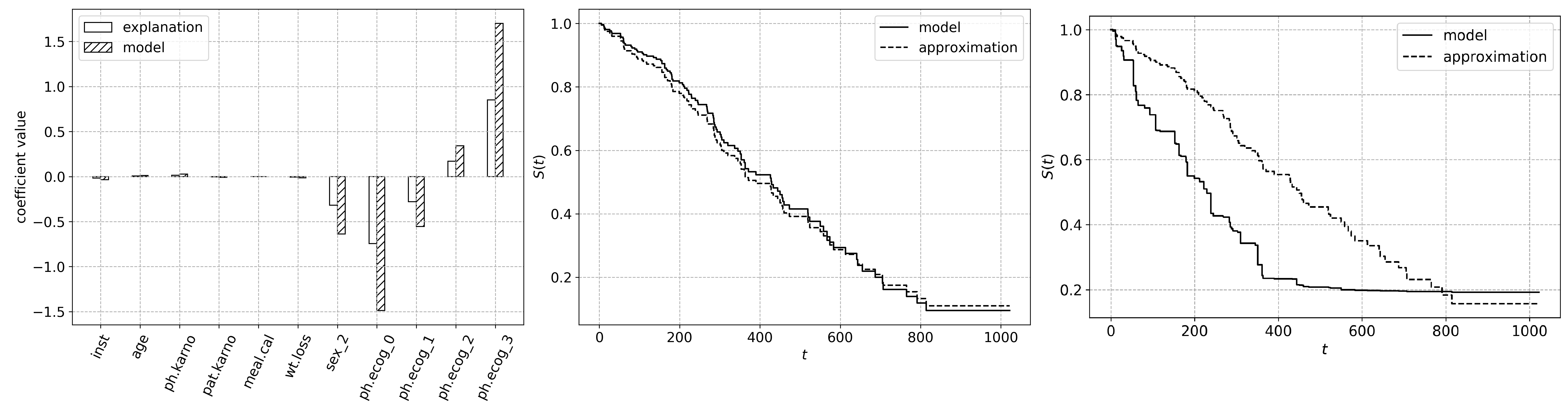}%
\caption{The mean approximation for the Cox model (the first and the second
picture) and the RSF (the third picture) trained on the LUNG dataset}%
\label{f:lung_cox_rsf}%
\end{center}
\end{figure}

\begin{figure}
[ptb]
\begin{center}
\includegraphics[
height=1.5333in,
width=5.5651in
]%
{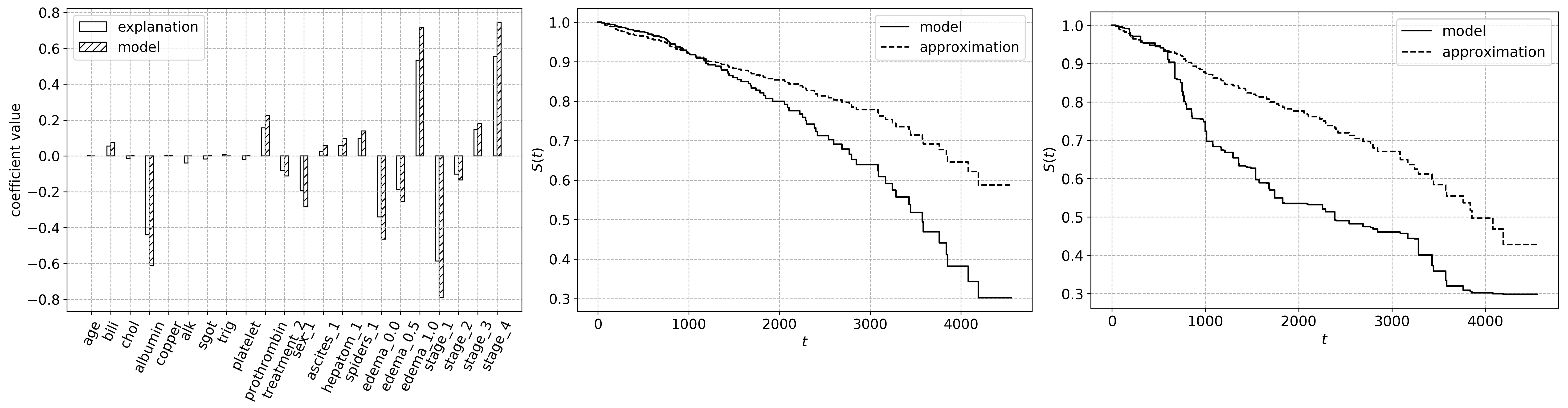}%
\caption{The mean approximation for the Cox model (the first and the second
picture) and the RSF (the third picture) trained on the PBC dataset}%
\label{f:pbc_cox_rsf}%
\end{center}
\end{figure}

Similar numerical results for the LUNG and PBC datasets are shown in Fig.
\ref{f:lung_cox_rsf} and Fig. \ref{f:pbc_cox_rsf}, respectively.%

%

%

\section{Conclusion}

A new modification of SurvLIME using the $L_{\infty}$-norm for computing the
distance between CHFs instead of the $L_{2}$-norm has been presented in the
paper. The basic idea behind both the methods is to approximate a survival
machine learning model at a point by the Cox proportional hazards model which
assumes a linear combination of the instance covariates. However, this idea is
differently implemented in SurvLIME and SurvLIME-Inf. SurvLIME-Inf extends the
set of explanation methods dealing with censored data in the framework of
survival analysis. Numerical experiments with synthetic and real datasets have
clearly illustrated accuracy and correctness of SurvLIME-Inf.

The main advantage of SurvLIME-Inf is that it uses the linear programming for
computing coefficients of the approximating Cox model. This peculiarity allows
us to develop new methods taking into account inaccuracy of training data,
possible imprecision of data. This is an interesting and important direction
for further research. Another problem, which can be solved by using
SurvLIME-Inf, is to explain machine learning survival models by using the Cox
model with time-dependent covariates. This is also an important direction for
further research.

\section*{Acknowledgement}

The reported study was funded by RFBR, project number 20-01-00154.

\bibliographystyle{plain}
\bibliography{Classif_bib,Deep_Forest,Explain,Explain_med,Medical,MYBIB,MYUSE,Robots,Survival_analysis}

\begin{thebibliography}{10}

\bibitem{Aas-etal-2019}
K.~Aas, M.~Jullum, and A.~Loland.
\newblock Explaining individual predictions when features are dependent: More
  accurate approximations to {S}hapley values.
\newblock arXiv:1903.10464, Mar 2019.

\bibitem{Adadi-Berrada-2018}
A.~Adadi and M.~Berrada.
\newblock Peeking inside the black-box: A survey on explainable artificial
  intelligence ({XAI}).
\newblock {\em IEEE Access}, 6:52138--52160, 2018.

\bibitem{Ahern-etal-2019}
I.~Ahern, A.~Noack, L.~Guzman-Nateras, D.~Dou, B.~Li, and J.~Huan.
\newblock Norm{L}ime: A new feature importance metric for explaining deep
  neural networks.
\newblock arXiv:1909.04200, Sep 2019.

\bibitem{Arrieta-etal-2019}
A.B. Arrieta, N.~Diaz-Rodriguez, J.~Del Ser, A.~Bennetot, S.~Tabik, A.~Barbado,
  S.~Garcia, S.~Gil-Lopez, D.~Molina, R.~Benjamins, R.~Chatila, and F.~Herrera.
\newblock Explainable artificial intelligence ({XAI}): Concepts, taxonomies,
  opportunities and challenges toward responsible {AI}.
\newblock arXiv:1910.10045, October 2019.

\bibitem{Arya-etal-2019}
V.~Arya, R.K.E. Bellamy, P.-Y. Chen, A.~Dhurandhar, M.~Hind, S.C. Hoffman,
  S.~Houde, Q.V. Liao, R.~Luss, A.~Mojsilovic, S.~Mourad, P.~Pedemonte,
  R.~Raghavendra, J.~Richards, P.~Sattigeri, K.~Shanmugam, M.~Singh, K.R.
  Varshney, D.~Wei, and Y.~Zhang.
\newblock One explanation does not fit all: A toolkit and taxonomy of {AI}
  explainability techniques.
\newblock arXiv:1909.03012, Sep 2019.

\bibitem{Barthe-etal-2005}
F.~Barthe, O.~Guedon, S.~Mendelson, and A.~Naor.
\newblock A probabilistic approach to the geometry of the l-ball.
\newblock {\em The Annals of Probability}, 33(2):480--513, 2005.

\bibitem{Bender-etal-2005}
R.~Bender, T.~Augustin, and M.~Blettner.
\newblock Generating survival times to simulate cox proportional hazards
  models.
\newblock {\em Statistics in Medicine}, 24(11):1713--1723, 2005.

\bibitem{Bou-Hamad-etal-2011}
I.~Bou-Hamad, D.~Larocque, and H.~Ben-Ameur.
\newblock A review of survival trees.
\newblock {\em Statistics Surveys}, 5:44--71, 2011.

\bibitem{Breiman-2001}
L.~Breiman.
\newblock Random forests.
\newblock {\em Machine learning}, 45(1):5--32, 2001.

\bibitem{Breslow-1974}
N.~Breslow.
\newblock Covariance analysis of censored survival data.
\newblock {\em Biometrics}, 30:89--99, 1974.

\bibitem{Buhrmester-etal-2019}
V.~Buhrmester, D.~Munch, and M.~Arens.
\newblock Analysis of explainers of black box deep neural networks for computer
  vision: A survey.
\newblock arXiv:1911.12116v1, November 2019.

\bibitem{Carvalho-etal-2019}
D.V. Carvalho, E.M. Pereira, and J.S. Cardoso.
\newblock Machine learning interpretability: A survey on methods and metrics.
\newblock {\em Electronics}, 8(832):1--34, 2019.

\bibitem{Cox-1972}
D.R. Cox.
\newblock Regression models and life-tables.
\newblock {\em Journal of the Royal Statistical Society, Series B
  (Methodological)}, 34(2):187--220, 1972.

\bibitem{Du-Liu-Hu-2019}
M.~Du, N.~Liu, and X.~Hu.
\newblock Techniques for interpretable machine learning.
\newblock arXiv:1808.00033, May 2019.

\bibitem{Efron-1977}
B.~Efron.
\newblock The efficiency of {C}ox likelihood function for censored data.
\newblock {\em Journal of the American Statistical Association}, 72:557--565,
  1977.

\bibitem{Faraggi-Simon-1995}
D.~Faraggi and R.~Simon.
\newblock A neural network model for survival data.
\newblock {\em Statistics in medicine}, 14(1):73--82, 1995.

\bibitem{Fleming-Harrington-1991}
T.R. Fleming and D.P. Harrington.
\newblock {\em Counting processes and survival aalysis}.
\newblock John Wiley \& Sons, Hoboken, NJ, USA, 1991.

\bibitem{Fong-Vedaldi-2019}
R.~Fong and A.~Vedaldi.
\newblock Explanations for attributing deep neural network predictions.
\newblock In {\em Explainable AI}, volume 11700 of {\em LNCS}, pages 149--167.
  Springer, Cham, 2019.

\bibitem{Fong-Vedaldi-2017}
R.C. Fong and A.~Vedaldi.
\newblock Interpretable explanations of black boxes by meaningful perturbation.
\newblock In {\em Proceedings of the IEEE International Conference on Computer
  Vision}, pages 3429--3437. IEEE, 2017.

\bibitem{Garreau-Luxburg-2020}
D.~Garreau and U.~von Luxburg.
\newblock Explaining the explainer: A first theoretical analysis of {LIME}.
\newblock arXiv:2001.03447, January 2020.

\bibitem{Guidotti-2019}
R.~Guidotti, A.~Monreale, S.~Ruggieri, F.~Turini, F.~Giannotti, and
  D.~Pedreschi.
\newblock A survey of methods for explaining black box models.
\newblock {\em ACM computing surveys}, 51(5):93, 2019.

\bibitem{Harman-Lacko-2010}
R.~Harman and V.~Lacko.
\newblock On decompositional algorithms for uniform sampling from n-spheres and
  n-balls.
\newblock {\em Journal of Multivariate Analysis}, 101:2297--2304, 2010.

\bibitem{Harrell-etal-1982}
F.~Harrell, R.~Califf, D.~Pryor, K.~Lee, and R.~Rosati.
\newblock Evaluating the yield of medical tests.
\newblock {\em Journal of the American Medical Association}, 247:2543--2546,
  1982.

\bibitem{Holzinger-etal-2017}
A.~Holzinger, C.~Biemann, C.S. Pattichis, and D.B. Kell.
\newblock What do we need to build explainable ai systems for the medical
  domain?
\newblock arXiv:1712.09923, Dec 2017.

\bibitem{Hosmer-Lemeshow-May-2008}
D.~Hosmer, S.~Lemeshow, and S.~May.
\newblock {\em Applied Survival Analysis: Regression Modeling of Time to Event
  Data}.
\newblock John Wiley \& Sons, New Jersey, 2008.

\bibitem{Hu-Chen-Nair-Sudjianto-2018}
L.~Hu, J.~Chen, V.N. Nair, and A.~Sudjianto.
\newblock Locally interpretable models and effects based on supervised
  partitioning ({LIME-SUP}).
\newblock arXiv:1806.00663, Jun 2018.

\bibitem{Huang-Yamada-etal-2020}
Q.~Huang, M.~Yamada, Y.~Tian, D.~Singh, D.~Yin, and Y.~Chang.
\newblock Graph{LIME}: Local interpretable model explanations for graph neural
  networks.
\newblock arXiv:2001.06216, January 2020.

\bibitem{Kalbfleisch-Prentice-1980}
J.~Kalbfleisch and R.~Prentice.
\newblock {\em The Statistical Analysis of Failure Time Data}.
\newblock John Wiley and Sons, New York, 1980.

\bibitem{Kim-etal-2012}
J.~Kim, I.~Sohn, S.-H. Jung, S.~Kim, and C.~Park.
\newblock Analysis of survival data with group lasso.
\newblock {\em Communications in Statistics - Simulation and Computation},
  41(9):1593--1605, 2012.

\bibitem{Kovalev-Utkin-Kasimov-20}
M.S. Kovalev, L.V. Utkin, and E.M. Kasimov.
\newblock Surv{LIME}: {A} method for explaining machine learning survival
  models.
\newblock arXiv:2003.08371, March 2020.

\bibitem{Lee-Zame-etal-2018}
C.~Lee, W.R. Zame, J.~Yoon, and M.~van~der Schaar.
\newblock Deephit: A deep learning approach to survival analysis with competing
  risks.
\newblock In {\em 32nd Association for the Advancement of Artificial
  Intelligence ( AAAI) Conference}, pages 1--8, 2018.

\bibitem{Li-Fan-Shi-Chou-19}
H.~Li, W.~Fan, S.~Shi, and Q.~Chou.
\newblock A modified lime and its application to explain service supply chain
  forecasting.
\newblock In {\em CCF International Conference on Natural Language Processing
  and Chinese Computing}, pages 637--644, Cham, 2019. Springer.

\bibitem{Loprinzi-etal-1994}
C.L. Loprinzi, J.A. Laurie, H.S. Wieand, J.E. Krook, P.J. Novotny, J.W. Kugler,
  J.~Bartel, M.~Law, M.~Bateman, and N.E. Klatt.
\newblock Prospective evaluation of prognostic variables from patient-completed
  questionnaires. north central cancer treatment group.
\newblock {\em Journal of Clinical Oncology}, 3(12):601--607, 1994.

\bibitem{Lundberg-Lee-2017}
S.M. Lundberg and S.-I. Lee.
\newblock A unified approach to interpreting model predictions.
\newblock In {\em Advances in Neural Information Processing Systems}, pages
  4765--4774, 2017.

\bibitem{Marosevic-1996}
T.~Marosevic.
\newblock A choice of norm in discrete approximation.
\newblock {\em Mathematical Communications}, 1(2):147--152, 1996.

\bibitem{May-etal-2004}
M.~May, P.~Royston, M.~Egger, A.C. Justice, and J.A.C. Sterne.
\newblock Development and validation of a prognostic model for survival time
  data: application to prognosis of {HIV} positive patients treated with
  antiretroviral therapy.
\newblock {\em Statistics in Medicine}, 23:2375--2398, 2004.

\bibitem{Mohseni-etal-2018}
S.~Mohseni, N.~Zarei, and E.D. Ragan.
\newblock A survey of evaluation methods and measures for interpretable machine
  learning.
\newblock arXiv:1811.11839, Dec 2018.

\bibitem{Molnar-2019}
C.~Molnar.
\newblock {\em Interpretable Machine Learning: A Guide for Making Black Box
  Models Explainable}.
\newblock Published online,
  https://christophm.github.io/interpretable-ml-book/, 2019.

\bibitem{Murdoch-etal-2019}
W.J. Murdoch, C.~Singh, K.~Kumbier, R.~Abbasi-Asl, and B.~Yua.
\newblock Interpretable machine learning: definitions, methods, and
  applications.
\newblock arXiv:1901.04592, Jan 2019.

\bibitem{Nezhad-etal-2018}
M.Z. Nezhad, N.~Sadati, K.~Yang, and D.~Zhu.
\newblock A deep active survival analysis approach for precision treatment
  recommendations: Application of prostate cancer.
\newblock arXiv:1804.03280v1, April 2018.

\bibitem{Petsiuk-etal-2018}
V.~Petsiuk, A.~Das, and K.~Saenko.
\newblock Rise: Randomized input sampling for explanation of black-box models.
\newblock arXiv:1806.07421, June 2018.

\bibitem{Rabold-etal-2019}
J.~Rabold, H.~Deininger, M.~Siebers, and U.~Schmid.
\newblock Enriching visual with verbal explanations for relational concepts:
  Combining {LIME} with {A}leph.
\newblock arXiv:1910.01837v1, October 2019.

\bibitem{Ramon-etal-2020}
Y.~Ramon, D.~Martens, F.~Provost, and T.~Evgeniou.
\newblock Counterfactual explanation algorithms for behavioral and textual
  data.
\newblock arXiv:1912.01819, December 2019.

\bibitem{Ribeiro-etal-2016}
M.T. Ribeiro, S.~Singh, and C.~Guestrin.
\newblock ``{W}hy should {I} trust {Y}ou?'' {E}xplaining the predictions of any
  classifier.
\newblock arXiv:1602.04938v3, Aug 2016.

\bibitem{Ribeiro-etal-2018}
M.T. Ribeiro, S.~Singh, and C.~Guestrin.
\newblock Anchors: High-precision model-agnostic explanations.
\newblock In {\em AAAI Conference on Artificial Intelligence}, pages
  1527--1535, 2018.

\bibitem{Rudin-2019}
C.~Rudin.
\newblock Stop explaining black box machine learning models for high stakes
  decisions and use interpretable models instead.
\newblock {\em Nature Machine Intelligence}, 1:206--215, 2019.

\bibitem{Shankaranarayana-Runje-2019}
S.M. Shankaranarayana and D.~Runje.
\newblock {ALIME}: Autoencoder based approach for local interpretability.
\newblock arXiv:1909.02437, Sep 2019.

\bibitem{Shi-Zhang-Fan-20}
S.~Shi, X.~Zhang, and W.~Fan.
\newblock A modified perturbed sampling method for local interpretable
  model-agnostic explanation.
\newblock arXiv:2002.07434.

\bibitem{Sim-Hartley-2006}
K.~Sim and R.~Hartley.
\newblock Removing outliers using the $l_\infty $ norm.
\newblock In {\em IEEE Computer Society Conference on Computer Vision and
  Pattern Recognition (CVPR'06)}, volume~1, pages 485--494, New York, NY, USA,
  2006.

\bibitem{Strumbel-Kononenko-2010}
E.~Strumbel and I.~Kononenko.
\newblock An efficient explanation of individual classifications using game
  theory.
\newblock {\em Journal of Machine Learning Research}, 11:1--18, 2010.

\bibitem{Tibshirani-1997}
R.~Tibshirani.
\newblock The lasso method for variable selection in the cox model.
\newblock {\em Statistics in medicine}, 16(4):385--395, 1997.

\bibitem{Tjoa-Guan-2019}
E.~Tjoa and C.~Guan.
\newblock A survey on explainable artificial intelligence ({XAI}): towards
  medical {XAI}.
\newblock arXiv:1907.07374v3, October 2019.

\bibitem{Vu-etal-2019}
M.N. Vu, T.D. Nguyen, N.~Phan, and M.T.~Thai R.~Gera.
\newblock Evaluating explainers via perturbation.
\newblock arXiv:1906.02032v1, Jun 2019.

\bibitem{Wachter-etal-2017}
S.~Wachter, B.~Mittelstadt, and C.~Russell.
\newblock Counterfactual explanations without opening the black box: Automated
  decisions and the {GPDR}.
\newblock {\em Harvard Journal of Law \& Technology}, 31:841--887, 2017.

\bibitem{Wang-Li-Reddy-2017}
P.~Wang, Y.~Li, and C.K. Reddy.
\newblock Machine learning for survival analysis: A survey.
\newblock arXiv:1708.04649, August 2017.

\bibitem{White-Garcez-2020}
A.~White and A.dA. Garcez.
\newblock Measurable counterfactual local explanations for any classifier.
\newblock arXiv:1908.03020v2, November 2019.

\bibitem{Zafar-Khan-2019}
M.R. Zafar and N.M. Khan.
\newblock {DLIME}: A deterministic local interpretable model-agnostic
  explanations approach for computer-aided diagnosis systems.
\newblock arXiv:1906.10263, Jun 2019.

\bibitem{Zhang-Lu-2007}
H.H. Zhang and W.~Lu.
\newblock Adaptive {L}asso for {C}ox's proportional hazards model.
\newblock {\em Biometrika}, 94(3):691--703, 2007.

\bibitem{Zhao-Feng-2019}
L.~Zhao and D.~Feng.
\newblock Dnnsurv: Deep neural networks for survival analysis using pseudo
  values.
\newblock arXiv:1908.02337v2, March 2020.

\end{thebibliography}

\end{document}